\documentclass{article}
\usepackage{coling2000}
\usepackage{epsfig}
\usepackage{rotating}
\usepackage{nobi}
\usepackage{nobiDG}

\begin{document}


\title{	
	The Use of Instrumentation in Grammar Engineering}

\author{Norbert Br\"{o}ker 
	\\ Eschenweg 3, 69231 Rauenberg }

\maketitle

\begin{abstract}
This paper explores the usefulness of a technique from software
engineering, code instrumentation, for the development of large-scale
natural language grammars. Information about the usage of grammar rules in
test and corpus sentences is used to improve grammar and testsuite, as well
as adapting a grammar to a specific genre. Results show that less than half
of a large-coverage grammar for German is actually tested by two large
testsuites, and that 10--30\% of testing time is redundant. This
methodology applied can be seen as a re-use of grammar writing knowledge
for testsuite compilation. The construction of genre-specific grammars
results in performance gains of a factor of four. 
\footnote{
The experiments reported here were conducted during my work at the Institut
f\"{u}r Maschinelle Sprachverarbeitung (IMS), Stuttgart University,
Germany. I'd like to thank Jonas Kuhn (IMS) and John Maxwell (Xerox PARC)
for their help in conducting these experiments.
}
\end{abstract}

\section{Introduction}
	\lbl{ch:intro}

The field of Computational Linguistics (CL) has both moved towards
applications and towards large data sets. These developments call for a
rigorous methodology for creating so-called lingware: linguistic data such
as lexica, grammars, tree-banks, as well as software processing it.
Experience from Software Engineering has shown that the earlier
deficiencies are detected, the less costly their correction is. Rather than
being a post-development effort, quality evaluation must be an integral
part of development to make the construction of lingware more efficient
(e.g., cf. \cite{EAGLES-EWG-PR.2} for a general evaluation framework and
\cite{Ciravegna1998} for the application of a particular software design
methodology to linguistic engineering).  This paper presents the adaptation
of a particular Software Engineering (SE) method, instrumentation, to
Grammar Engineering (GE). Instrumentation allows to determine which test
item exercises a certain piece of (software or grammar) code.

The paper first describes the use of instrumentation in SE, then discusses
possible realizations in unification grammars, and finally presents two
classes of applications.

\section{Software Instrumentation}
	\lbl{ch:se}

Systematic software testing requires a match between the test subject
(module or complete system) and a test suite (collection of test items,
i.e., sample input). This match is usually computed as the percentage of
code items exercised by the test suite.

Depending on the definition of a code item, various measures are employed,
for example (cf. \cite{Hetzel1988} and \cite[Appendix B]{EAGLES-EWG-PR.2}
for overviews):

\begin{description}
\item[statement coverage] percentage of single statements exercised
\item[branch coverage] percentage of arcs exercised in control flow graph;
subsumes statement coverage
\item[path coverage] percentage of paths exercised from start to end in
control flow graph; subsumes branch coverage; impractical due to large
(often infinite) number of paths
\item[condition coverage] percentage of (simple or aggregate) conditions
evaluated to both true and false (on different test items)
\end{description}

Testsuites are constructed to maximize the targeted measure. A test run
yields information about the code items not exercised, allowing the
improvement of the testsuite. 

The measures are automatically obtained by instrumentation: The test
subject is extended by code which records the code items exercised during
processing. After processing the testsuite, the records are used to compute
the measures.

\section{Grammar Instrumentation}
	\lbl{ch:ge}

Measures from SE cannot simply be transferred to unification grammars,
because the structure of (imperative) programs is different from
that of (declarative) grammars. Nevertheless, the structure of a grammar
(formalism) allows to define measures very similar to those employed in SE.

\begin{description}
\item[constraint coverage] is the quotient 
\[
T_{\mbox{con}}=\frac{\mbox{\# constraints exercised}}{\mbox{\# constraint in grammar}} 
\]
where a constraint may be either a phrase-structure or an equational
constraint, depending on the formalism. 

\item[disjunction coverage] is the quotient 
\[
T_{\mbox{dis}}=\frac{\mbox{\# disjunctions covered}}{\mbox{\# disjunctions in grammar}} 
\]
where a disjunction is considered covered when all its alternative
disjuncts have been separately exercised. It encompasses constraint
coverage. Optional constituents and equations have to be treated as a
disjunction of the constraint and an empty constraint (cf.
Fig.\ref{fig:instr-example} for an example).

\item[interaction coverage] is the quotient 
\[
T_{\mbox{int}}=\frac{\mbox{\# disjunct combinations exercised}}{\mbox{\# legal disjunct combinations}} 
\]
where a disjunct combination is a complete set of choices in the
disjunctions which yields a well-formed grammatical structure.

As with path coverage, the set of legal disjunct combination typically is
infinite due to recursion. A solution from SE is to restrict the use of
recursive rules to a fixed number of cases, for example not using the rule
at all, and using it only once.
\end{description}

The goal of instrumentation is to obtain information about which test cases
exercise which grammar constraints. One way to record this information is
to extend the parsing algorithm. Another way is to use the grammar
formalism itself to identify the disjuncts. Depending on the expressivity
of the formalism used, the following possibilities exist:

\begin{description}
\item[atomic features] Assuming a unique numbering of disjuncts, an
annotation of the form {\tt DISJUNCT-nn~=~+} can be used for marking. To
determine whether a certain disjunct was used in constructing a solution,
one only needs to check whether the associated feature occurs (at some
level of embedding) in the solution.

\item[set-valued features] If set-valued features are available, one can
use a set-valued feature {\tt DISJUNCTS} to collect atomic symbols
representing one disjunct each: {\tt DISJUNCT-nn~$\in$~DISJUNCTS}, which
might ease the collection of exercised disjuncts. 

\item[multiset of symbols] To recover the number of times a disjunct is
used, one needs to leave the unification paradigm, because it is very
difficult to count with unification grammars. 
The Xerox Linguistic Environment used here (XLE; cf. {\tt
www.parc.xerox.com/istl/groups/nltt/pargram} and \cite{KaplanNewman1997})
provides for a multiset of symbols to be associated with each complete
structural analysis: Following the LFG spirit of different projections, it
defines a projection of symbolic marks which is formally equivalent to a
multiset of symbols (cf. \cite{Frank1998} for an introduction and several
applications). Thus, one may recover the set of all disjuncts used from
each analysis, together with their frequency.
\end{description}

Consider the LFG grammar rule in Fig.\ref{fig:rule}.%
\footnote{
Although the sample rules are in the format of LFG, nothing of the
methodology relies on the choice of linguistic or computational paradigm.
The notation: \texttt{?}/\texttt{*}/\texttt{+} represent
optionality/iteration including/excluding zero occurrences on categories.
\texttt{e} represents the empty string. Annotations to a category specify
equality (\texttt{=}) or set membership ($\in$) of feature values, or
non-existence of features ($\neg$); they are terminated by a semicolon
(\texttt{;}). Disjunctions are given in braces
(\texttt{\{...\|...\}}). $\uparrow$ ($\downarrow$) are metavariables
representing the feature structure corresponding to the mother (daughter)
of the rule. $o*$ (for optimality) represents the sentence's multi-set
valued symbolic projection.  Comments are enclosed in quotation marks
(\texttt{"..."}). Cf.  \cite{Kaplan+Bresnan1982} for an introduction to LFG
notation.  
}
Constraint coverage would require test items such that every category in
the \texttt{VP} is exercised; a sequence of \texttt{V NP PP} would suffice
for this measure. Disjunction coverage also requires to take the empty
disjuncts into account: \texttt{NP} and \texttt{PP} are optional, so that
four items are needed to achieve full disjunction coverage on the phrase
structure part of the rule. Due to the disjunction in the \texttt{PP}
annotation, two more test items are required to achieve full disjunction
coverage on the complete rule. Fig.\ref{fig:instr-example} shows the rule
from Fig.\ref{fig:rule} with instrumentation.

\begin{figure}[t]
\small \centering
\begin{tabular}{llcl}
VP \prod	& V	& \mcl{2}{$\downarrow = \uparrow$;} \\
		& NP?	& \mcl{2}{$\downarrow = (\uparrow \mbox{OBJ})$;} \\
		& PP*	& \{	& $\downarrow = (\uparrow \mbox{OBL})$; \\
		&	& \|	& $\downarrow \in (\uparrow \mbox{ADJUNCT})$; \}.
\end{tabular}
\caption{Sample Rule 
	\label{fig:rule}}
\end{figure}

\begin{figure}[t]
\small \centering
\begin{tabular}{llcl}
VP \prod	& V	& \mcl{2}{$\downarrow = \uparrow$;} \\
		& \{ e	& \mcl{2}{DISJUNCT-001 $\in o*$;} \\
		& \| NP	& \mcl{2}{$\downarrow = (\uparrow \mbox{OBJ})$} \\
		&	& \mcl{2}{DISJUNCT-002 $\in o*$; \}} \\
		& \{ e	& \mcl{2}{DISJUNCT-003 $\in o*$;} \\
		& \| PP+& \{	& $\downarrow = (\uparrow \mbox{OBL})$ \\
		&	&	& DISJUNCT-004 $\in o*$; \\
		&	& \|	& $\downarrow \in (\uparrow \mbox{ADJUNCT})$ \\
		&	&	& DISJUNCT-005 $\in o*$;\} \}. \\
\end{tabular}
\caption{Instrumented rule
	\label{fig:instr-example}}
\end{figure}

\section{Grammar and Testsuite Improvement}
	\lbl{ch:testsuite}

Traditionally, a testsuite is used to improve (or maintain) a grammar's
quality (in terms of coverage and overgeneration). Using instrumentation,
one may extend this usage by looking for sources of overgeneration
(cf. Sec.\ref{ch:negative}), and may also improve the quality of the
testsuite, in terms of coverage (cf. Sec.\ref{ch:suite-complete}) and
economy (cf. Sec.\ref{ch:economic}).

Complementing other work on testsuite construction (cf.
Sec.\ref{ch:compare-suites}), I will assume that a grammar is already
available, and that a testsuite has to be constructed or extended. While
one may argue that grammar and testsuite should be developed in parallel,
such that the coding of a new grammar disjunct is accompanied by the
addition of suitable test cases, and vice versa, this is seldom the case.
Apart from the existence of grammars which lack a testsuite, there is the
more principled obstacle of the evolution of the grammar, leading to states
where previously necessary rules silently loose their usefulness, because
their function is taken over by some other rules, structured
differently. This is detectable by instrumentation, as discussed in
Sec.\ref{ch:suite-complete}.

On the other hand, once there is a testsuite, it has to be used
economically, avoiding redundant tests. Sec.\ref{ch:economic} shows that
there are different levels of redundancy in a testsuite, dependent on the
specific grammar used. Reduction of this redundancy can speed up the test
activity, and give a clearer picture of the grammar's performance.

\subsection{Testsuite Completeness}
	\lbl{ch:suite-complete}

If the disjunction coverage of a testsuite is 1 for some grammar, the
testsuite is \emph{complete} w.r.t. this grammar. Such a testsuite can
reliably be used to monitor changes in the grammar: Any reduction in the
grammar's coverage will show up in the failure of some test case (for
negative test cases, cf. Sec.\ref{ch:negative}).

If the testsuite is not complete, instrumentation can identify disjuncts
which are not exercised. These might be either (i) appropriate, but
untested, disjuncts calling for the addition of a test case, or (ii)
inappropriate disjuncts, for which a grammatical test case exercising them
cannot be constructed.

Experiments were based on a large German LFG grammar developed at the IMS
(cf. {\tt www.ims.uni-stuttgart.de/projekte/pargram} and
\cite{KuhnEckleRohrer1998,KuhnRohrer1997}). We found that a testsuite of
1787 items collected to support grammar development only exercised 1456 out
of 3730 grammar disjuncts, yielding $T_{dis} = 0.39$. The TSNLP testsuite
containing 1093 items exercised only 1081 disjuncts, yielding $T_{dis} =
0.28$).%
\footnote{%
There are, of course, unparsed but grammatical test cases in both
testsuites, which have not been taken into account in these figures. This
explains the difference to the overall number of 1582 items in the German
TSNLP testsuite. 
} 
Fig.\ref{fig:gap} shows an example of a gap in our testsuite (there are no
examples of circumpositions), while Fig.\ref{fig:inappropriate} shows an
inapproppriate disjunct thus discovered (the category ADVadj has been
eliminated in the lexicon, but not in all rules). 

\begin{figure}[t]
\small \centering
\begin{tabular}{lll}
PPstd \prod	& Pprae		& $\downarrow = \uparrow$;\\
		& NPstd		& $\downarrow = (\uparrow \mbox{OBJ})$; \\
		& \{ e		& DISJUNCT-011 $\in o*$; \\
		& \| Pcircum	& $\downarrow = \uparrow$;\\
		&		& DISJUNCT-012 $\in o*$ \\
		&		& "unused disjunct"; \} \\
\end{tabular}
\caption{Appropriate untested disjunct}
	\label{fig:gap}
\end{figure}

\begin{figure}[t]
\small \centering
\begin{tabular}{lcll}
ADVP \prod	& \{	& \{ e		& DISJUNCT-021 $\in o*$;\\
		&	& \| ADVadj	& $\downarrow = \uparrow$ \\
		&	&		& DISJUNCT-022 $\in o*$ \\
		&	&		& "unused disjunct"; \} \\
		& 	& ADVstd	& $\downarrow = \uparrow$ \\
		&	&		& DISJUNCT-023 $\in o*$ \\
		&	&		& "unused disjunct"; \} \\
		& \| 	& ... \}. \\
\end{tabular}
\caption{Inappropriate disjunct
	\label{fig:inappropriate}}
\end{figure}


\subsection{Testsuite Economy}
	\lbl{ch:economic}

Besides being complete, a testsuite must be economical, i.e., contain as
few items as possible. Instrumentation can identify redundant test cases,
where redundancy can be defined in three ways:

\begin{description}
\item[similarity] There is a set of other test cases which jointly exercise
all disjunct which the test case under consideration exercises.

\item[equivalence] There is a single test case which exercises exactly the
same combination(s) of disjuncts.

\item[strict equivalence] There is a single test case which is equivalent
to and, additionally, exercises the disjuncts exactly as often as, the test
case under consideration.
\end{description}

Fig.\ref{fig:equivs} shows equivalent test cases found in our testsuite:
Example 1 illustrates the distinction between equivalence and strict
equivalence; the test cases contain different numbers of attributive
adjectives. Example 2 shows that our grammar does not make any distinction
between adverbial usage and secondary (subject or object) predication.

\begin{figure}[t]
\small \centering
\begin{tabular}{ll}
1	& ein guter alter Wein \\
	& ein guter alter trockener Wein \\
	& `\emph{a good old (dry) wine}' \\
2	& Er i\ss t das Schnitzel roh. \\
	& Er i\ss t das Schnitzel nackt. \\
	& Er i\ss t das Schnitzel schnell. \\
	& `\emph{He eats the schnitzel naked/raw/quickly.}' \\
\end{tabular}
\caption{Sets of equivalent test cases
	\label{fig:equivs}}
\end{figure}

The achievable reduction in size and processing time is shown in
Table~\ref{tbl:reduced-suites}, which contains measurements for a test run
containing only the parseable test cases, one without equivalent test cases
(for every set of equivalent test cases, one was arbitrarily selected), and
one without similar test cases. The last was constructed using a simple
heuristic: Starting with the sentence exercising the most disjuncts,
working towards sentences relying on fewer disjuncts, a sentence was
selected only if it exercised a disjunct which no previously selected
sentence exercised. Assuming that a disjunct working correctly once will
work correctly more than once, we did not consider strict equivalence.

\begin{table}
\small \centering
\begin{tabular}{@{}l@{}rrrr@{}}
			& \shortstack{test \\ cases}
				& \shortstack{relative \\ size}
					& \shortstack{runtime \\ (sec)} 
						& \shortstack{relative \\ runtime} \\
\hline
\multicolumn{5}{c}{TSNLP testsuite} \\
\hline
parseable		& 1093	& 100\%	& 1537	& 100\%	\\
no equivalents		& 783	& 71\%	& 665.3	& 43\%	\\
no similar cases	& 214	& 19\%	& 128.5	& 8\%	\\
\hline
\multicolumn{5}{c}{local testsuite} \\
\hline
parseable		& 1787	& 100\%	& 1213	& 100\%	\\
no equivalents		& 1600	& 89\%	& 899.5	& 74\%	\\
no similar cases	& 331	& 18\%	& 175.0	& 14\%	\\
\hline
\end{tabular}
\caption{Reduction of Testsuites
	\label{tbl:reduced-suites}}
\end{table}

We envisage the following use of this redundancy detection: There clearly
are linguistic reasons to distinguish all test cases in example 2, so they
cannot simply be deleted from the testsuite. Rather, their equivalence
indicates that the grammar is not yet perfect (or never will be, if it
remains purely syntactic). Such equivalences could be interpreted as a
reminder which linguistic distinctions need to be incorporated into the
grammar. Thus, this level of redundancy may drive your grammar development
agenda. The level of equivalence can be taken as a limited interaction
test: These test cases represent one complete selection of grammar
disjuncts, and (given the grammar) there is nothing we can gain by checking
a test case if an equivalent one was tested. Thus, this level of redundancy
may be used for ensuring the quality of grammar changes prior to their
incorporation into the production version of the grammar. The level of
similarity contains much less test cases, and does not test any
(systematic) interaction between disjuncts. Thus, it may be used during
development as a quick rule-of-thumb procedure detecting serious errors
only.

\subsection{Sources of Overgeneration}
	\lbl{ch:negative}

To control overgeneration, appropriately marked ungrammatical sentences are
important in every testsuite. Instrumentation as proposed here only looks
at successful parses, but can still be applied in this context: If an
ungrammatical test case receives an analysis, instrumentation informs us
about the disjuncts used in the incorrect analysis. One of these disjuncts
must be incorrect, or the sentence would not have received a solution. We
exploit this information by accumulation across the entire test suite,
looking for disjuncts that appear in unusually high proportion in parseable
ungrammatical test cases.

In this manner, six grammar disjuncts are singled out by the parseable
ungrammatical test cases in the TSNLP testsuite. The most prominent
disjunct appears in 26 sentences (listed in Fig.\ref{fig:negative}), of
which the top left group is indeed grammatical and the rest fall into two
classes: A partial VP with object NP, interpreted as an imperative
sentence (bottom left), and a weird interaction with the tokenizer
incorrectly handling capitalization (right group).

\begin{figure}[t]
\small
\begin{tabular}{ll}
Der Test f\"{a}llt leicht. & Dieselbe schlafen. \\
Die schlafen.			& Das schlafen. \\     
				& Eines schlafen. \\   
Man schlafen. 			& Jede schlafen. \\    
Dieser schlafen. 		& Dieses schlafen. \\  
Ich schlafen. 			& Eine schlafen. \\    
Der schlafen. 			& Meins schlafen. \\   
Jeder schlafen.			& Dasjenige schlafen. \\
Derjenige schlafen. 		& Jedes schlafen. \\
Jener schlafen. 		& Diejenige schlafen. \\
Keiner schlafen. 		& Jenes schlafen. \\
Derselbe schlafen. 		& Keines schlafen. \\  
Er schlafen. 			& Dasselbe schlafen. \\
Irgendjemand schlafen. 		\\
\end{tabular}
\caption{Sentences relying on suspicious disjunct
	\label{fig:negative}}
\end{figure}

Far from being conclusive, the similarity of these sentences derived from a
suspicious grammar disjunct, and the clear relation of the sentences to
only two exactly specifiable grammar errors make it plausible that this
approach is very promising in detecting the sources of overgeneration.

\subsection{Other Approaches to Testsuite Construction}
	\lbl{ch:compare-suites}

The delicacy of testsuite construction is acknowledged in
\cite[p.37]{EAGLES-EWG-PR.2}. Although there are a number of efforts to
construct reusable testsuites, none has to my knowledge explored how
existing grammars can be exploited.

Starting with \cite{Flickinger1987}, testsuites have been drawn up from a
linguistic viewpoint, \emph{informed by [the] study of linguistics and
[reflecting] the grammatical issues that linguists have concerned
themselves with} \cite[p.4]{Flickinger1987}. Although the question is not
explicitly addressed in \cite{TSNLP-WP2.2}, all the testsuites reviewed
there also seem to follow the same methodology. The TSNLP project
\cite{Lehmann1996} and its successor DiET \cite{Netter1998}, which built
large multilingual testsuites, likewise fall into this category.

The use of corpora (with various levels of annotation) has been studied,
but the recommendations are that much manual work is required to turn
corpus examples into test cases (e.g., \cite{TSNLP-WP5.2}). The reason
given is that corpus sentences neither contain linguistic phenomena in
isolation, nor do they contain systematic variation. Corpora thus are used
only as an inspiration.

\cite{Oepen1998} stress the interdependence between application and
testsuite, but don't comment on the relation between grammar and
testsuite.

\section{Genre Adaptation}
	\lbl{ch:genre}

A different application of instrumentation is the tailoring of a general
grammar to specific genres.  All-purpose grammars are plagued by lexical
and structural ambiguity that leads to overly long runtimes. If this
ambiguity could be limited, parsing efficiency would improve. Instrumenting
a general grammar allows to automatically derive specialized subgrammars
based on sample corpora. This setup has several advantages: The larger the
overlap between genres, the larger the portion of grammar development work
that can be recycled.  The all-purpose grammar is linguistically more
interesting, because it requires an integrated concept, as opposed to
several separate genre-specific grammars.

I will discuss two ways of improving the efficiency of parsing a
sublanguage, given an all-purpose unification grammar. The first consists
in deleting unused disjuncts, while the second uses a staged parsing
process. The experiments are only sketched, to indicate the applicability
of the instrumentation technique, and not to directly compete with other
proposals on grammar specialization. For example, the work reported in
\cite{Rayner1994,Samuelsson1994} differs from the one presented
below in several aspects: They induce a grammar from a treebank, while I
propose to annotate the grammar based on all solutions it produces. No
criteria for tree decomposition and category specialization are needed
here, and the standard parsing algorithm can be used. On the other hand,
the efficiency gains are not as big as those reported by \cite{Rayner1994}.

\subsection{Restricting the Grammar}

Given a large sample of a genre, instrumentation allows you to determine
the likely constructions of that genre. Eliminating unused disjuncts allows
faster parsing due to a smaller grammar. An experiment was conducted with
several corpora as detailed in Table~\ref{tbl:corpora}. There was some
effort to cover the corpus HC-DE, but no grammar development based on the
other corpora. The NEWS-SC corpus is part the corpus of verb-final
sentences used by \cite{Beil1999}.

\begin{table}
\small \centering
\begin{tabular}{@{}llc@{}}
Descriptor    & Content					& Coverage \\
\hline						  
HC-DE         & Copier/Printer User Manual		& 89\% \\
WHB           & Car Maintenance Instructions		& 76\% \\
NEWS          & News (5-30 words per sentence)		& 42\% \\
NEWS-SC       & Verb-final subclauses from News		& 75\%\\
\end{tabular}
\caption{Corpora used for adaptation
	\label{tbl:corpora}}
\end{table}

A training set of 1000 sentences from each corpus was parsed with an
instrumented \emph{base grammar}. From the parsing results, the exercised
grammar disjuncts were extracted and used to construct a corpus-specific
\emph{reduced grammar}. The reduced grammars were then used to parse a test
set of another 1000 sentences from each corpus.
Table~\ref{tbl:performance} shows the performance improvement on the
corpora: It gives the size of the grammars in terms of the number of rules
(with regular expression right-hand sides and feature annotation), the
number of arcs (corresponding to unary or binary rules with disjunctive
feature annotation), and the number of disjuncts (unary or binary rules
with unique feature annotation). The number of mismatches counts the
sentences for which the solution(s) obtained differed from those obtained
with the base grammar, while the number of additions counts the sentences
which did not receive a parse with the base grammar due to resource
limitations (runtime or memory), but received one with the reduced
grammar. The other columns give timings to process the total corpus, and
the longest and average processing time per sentence; time is in seconds.
The last column gives the average number of solutions per sentence.

\begin{table*}
\small\centering
\begin{tabular}{l|rrr|rr|rrr|}
			& \begin{turn}{90}
			  \# of rules
			  \end{turn}
				& \begin{turn}{90}
				  \# of arcs	
				  \end{turn}
					& \begin{turn}{90}
					  \shortstack{\# of \\ disjuncts}
					  \end{turn}
						& \begin{turn}{90}
						  \shortstack{\# of \\ mismatches}
						  \end{turn}
							& \begin{turn}{90}
							  \shortstack{\# of \\ additions}
							  \end{turn}
								& \begin{turn}{90}
								  total time
								  \end{turn}
									& \begin{turn}{90}
									  \shortstack{max. time \\ per sentence}
									  \end{turn}
										& \begin{turn}{90}
										  \shortstack{avg. time \\ per sentence}
										  \end{turn} \\
\hline
\hline
Corpus HC-DE \\
base grammar		& 185	& 3669	& 11564	& n/a	& n/a	&7692.4	& $>$300	& 7.1\\
reduced grammar	(938)	& 112	& 960	& 3739	& 0	& 1	&2089.4	& 162.7	& 1.9 \\
\hline
Corpus WHB \\
base grammar		& 195	& 3728	& 11606	& n/a	&	& 1428.9& $>$300.3	& 1.5 \\
reduced grammar	(559)	& 	& 534	& 3072	& 1	&	& 444.2	& 11.3	& 0.4 \\
\end{tabular}
\caption{Performance of reduced grammars
	\label{tbl:performance}}
\end{table*}

\begin{figure*}
\epsfig{file=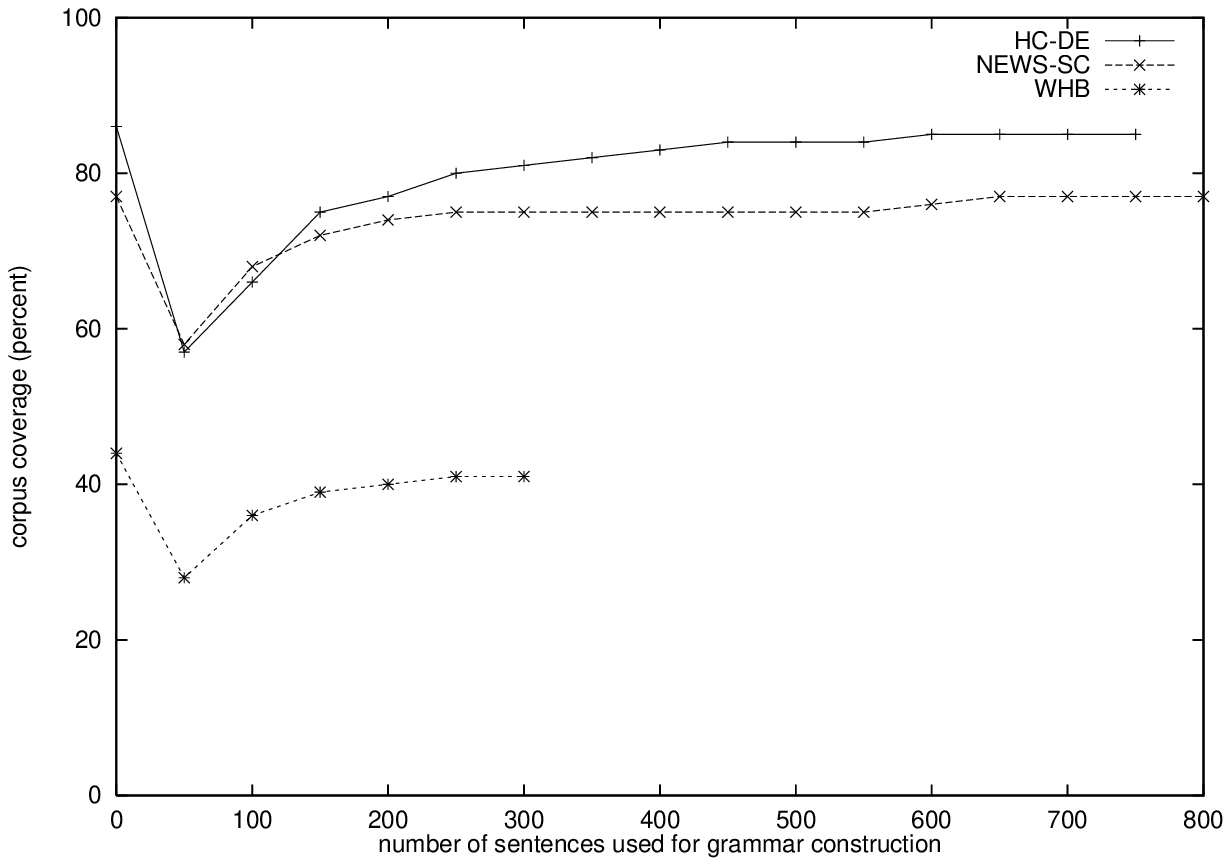,width=0.49\textwidth}
\hfill
\epsfig{file=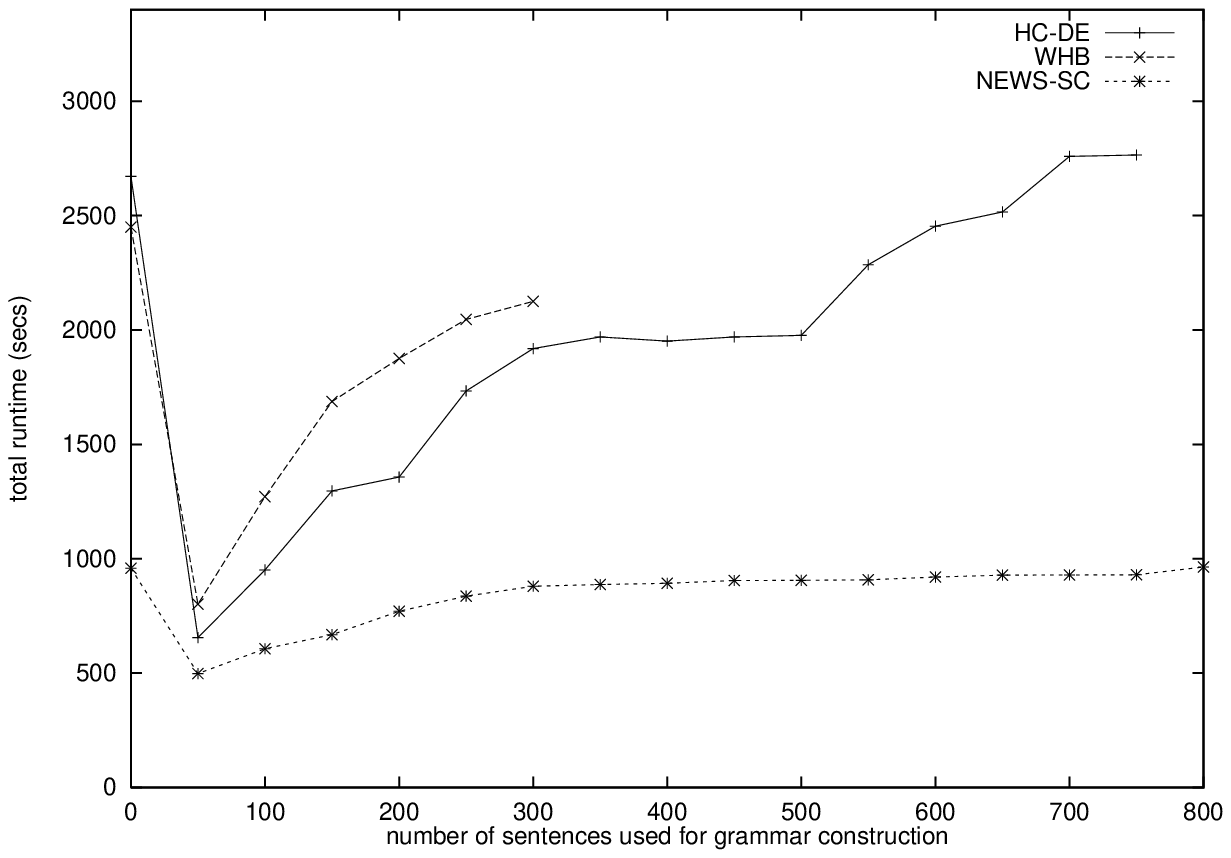,width=0.49\textwidth}
\caption{Performance of fragment grammars
	\label{fig:fragments}}
\end{figure*}

Due to the sampling of a genre, the grammars obtained can only be
approximate. To determine the relation of the sample size to the quality
of the grammar obtained, the coverage of random \emph{fragment grammars}
was measured in the following way: Randomly select a number of sentences
from the total corpus, construct (in the same way as described above for
the reduced grammar) a fragment grammar, and determine its coverage on the
test set from the respective corpus. The graphs in Fig.\ref{fig:fragments}
show how the coverage and runtime relate to the number of sentences on
which the fragment grammars are based. The leftmost data point (x value 0)
describes the performance of the reduced grammar on the training set, while
the rightmost data point describes its performance on the test set. The
data points in between represent fragment grammars based on as many
sentences as given by the x axis value. 

The results reported here represent the minimal performance gain due to the
fact that the construction of reduced and fragment grammars are not based
on the correct solutions for the training sentences, but rather on all
solutions produced by the base grammar. The construction of a large-scale
treebank with manually verified solutions is under way but has not yet
progressed far enough to serve as input for this experiment. Even with this
systematic, but curable error, the reduction reduces overall processing by
a factor of four. The number of solutions is constant because only unused
disjuncts are eliminated; this will change if the treebank solutions are
used to construct the reduced grammar.

\subsection{Staged Parsing}

Even eliminating only unlikely disjuncts necessarily reduces the coverage
of the grammar. A sequence of parsing stages allows one to profit from a
small and fast grammar as well as from a large and slow one. Staged parsing
applies different grammars one after the other to the input, until one
yields a solution, which terminates the process. In our case, a grammar of
stage $n+1$ includes the grammar of stage $n$, but this need not be the
case in general.

To reduce the variability for an experiment, I assume three stages: The
first includes frequently used disjuncts, the second all used disjuncts,
and the third all grammar disjuncts. This ensures the same coverage as the
base grammar, but allows to focus on frequent constructions in the first
parsing stage. The procedure is similar as before: From the solutions of a
training set, a \emph{staged grammar} is constructed. Currently,
experiments are performed to determine a useful definition of `frequently
used'.  Independent from the actual performance gains finally obtained, the
application of instrumentation allows a systematic exploration of the
possible configurations.

\subsection{Other approaches to grammar adaptation}
	\lbl{ch:comp-adapt}

\cite{Rayner1994,Rayner1996,Samuelsson1994} present a grammar
specialization technique for unification grammars. From a treebank of the
sublanguage, they induce a specialized grammar using fewer \emph{macro
rules} which correspond to the application of several original rules.  They
report an average speed-up of 55 for only the parsing phase (taking lexical
lookup into account, the speed-up factor was only 6--10). Due to the
derivation of the grammar from a corpus sample, they observed a decrease in
recall of 7.3\% and an increase of precision of 1.6\%. The differences to
the approach described here are clear: Starting from the grammar, rather
than from a treebank, we annotate the rules, rather than inducing them from
scratch. We do not need criteria for tree decomposition and category
specialization, and we can use the standard parsing algorithm.  On the
other hand, the efficiency gains are not as big as those reported by
\cite{Rayner1996} (but note that we cannot measure parsing times alone, so
we need to compare to their speed-up factor of 10). And we did not (yet)
start from a treebank, but from the raw set of solutions.

\section{Conclusion}
	\lbl{ch:conclu}

I have presented the adaptation of code instrumentation to Grammar
Engineering, discussing measures and implementations, and sketching several
applications together with preliminary results.  

The main application is to improve grammar and testsuite by exploring the
relation between both of them. Viewed this way, testsuite writing can
benefit from grammar development because both describe the syntactic
constructions of a natural language. Testsuites systematically list these
constructions, while grammars give generative procedures to construct them.
Since there are currently many more grammars than testsuites, we may re-use
the work that has gone into the grammars for the improvement of testsuites.

Other applications of instrumentation are possible; genre adaptation was
discussed in some depth. On a more general level, one may ask whether other
methods from SE may fruitfully apply to GE as well, possibly in modified
form. For example, the static analysis of programs, e.g.,  detection of
unreachable code, could also be applied for grammar development to detect
unusable rules.

\bibliographystyle{coling2000}

\begin{thebibliography}{}

\bibitem[\protect\citename{Balkan and Fouvry}1995]{TSNLP-WP5.2}
L.~Balkan and F.~Fouvry.
\newblock 1995.
\newblock {\em Corpus-based test suite generation}.
\newblock TSNLP-WP 2.2, University of Essex.

\bibitem[\protect\citename{Beil \bgroup et al.\egroup }1999]{Beil1999}
F.~Beil, G.~Carroll, D.~Prescher, S.~Riezler, and M.~Rooth.
\newblock 1999.
\newblock Inside-outside estimation of a lexicalized {PCFG} for german.
\newblock In {\em Proc. 37th Annual Meeting of the ACL}. Maryland.

\bibitem[\protect\citename{Ciravegna \bgroup et al.\egroup
  }1998]{Ciravegna1998}
F.~Ciravegna, A.~Lavelli, D.~Petrelli, and F.~Pianesi.
\newblock 1998.
\newblock Developing language reesources and applications with {GEPPETTO}.
\newblock In {\em Proc. 1st Int'l Conf. on Language Resources and Evaluation},
  pages 619--625. Granada/Spain, 28-30 May 1998.

\bibitem[\protect\citename{EAGLES}1996]{EAGLES-EWG-PR.2}
EAGLES.
\newblock 1996.
\newblock {\em Evaluation of Natural Language Processing Systems}.
\newblock Final Report EAG-EWG-PR.2.

\bibitem[\protect\citename{Estival \bgroup et al.\egroup }1994]{TSNLP-WP2.2}
D.~Estival, K.~Falkedal, S.~Lehmann, L.~Balkan, S.~Meijer, D.~Arnold,
  S.~Regnier-Prost, E.~Dauphin, K.~Netter, and S.~Oepen.
\newblock 1994.
\newblock {\em Test Suite Design --- Annotation Scheme}.
\newblock Number D-WP2.2.

\bibitem[\protect\citename{Flickinger \bgroup et al.\egroup
  }1987]{Flickinger1987}
D.~Flickinger, J.~Nerbonne, I.~Sag, and T.~Wasow.
\newblock 1987.
\newblock {\em Toward Evaluation of NLP Systems}.
\newblock Hewlett-Packard Laboratories, Palo Alto/CA.

\bibitem[\protect\citename{Frank \bgroup et al.\egroup }1998]{Frank1998}
A.~Frank, T.H. King, J.~Kuhn, and J.~Maxwell.
\newblock 1998.
\newblock Optimality theory style constraint ranking in large-scale {LFG}
  grammar.
\newblock In {\em Proc. of the LFG98 Conference}. Brisbane/AUS, Aug 1998, CSLI
  Online Publications.

\bibitem[\protect\citename{Hetzel}1988]{Hetzel1988}
W.C. Hetzel.
\newblock 1988.
\newblock {\em The complete guide to software testing}.
\newblock QED Information Sciences, Inc. Wellesley/MA 02181.

\bibitem[\protect\citename{Kaplan and Bresnan}1982]{Kaplan+Bresnan1982}
R.M. Kaplan and J.~Bresnan.
\newblock 1982.
\newblock Lexical-functional grammar: A formal system for grammatical
  representation.
\newblock In J.~Bresnan and R.M. Kaplan, editors, {\em The Mental
  Representation of Grammatical Relations}, pages 173--281. Cambridge, MA: MIT
  Press.

\bibitem[\protect\citename{Kaplan and Newman}1997]{KaplanNewman1997}
R.~Kaplan and P.~Newman.
\newblock 1997.
\newblock Lexical resource reconciliation in the {Xerox Linguistic
  Environment}.
\newblock In D.~Estival, A.~Lavelli, K.~Netter, and F.~Pianesi, editors, {\em
  Workshop ``Computational Environments for Grammar Development and Linguistic
  Engineering''}, pages 54--61. Madrid.

\bibitem[\protect\citename{Kuhn and Rohrer}1997]{KuhnRohrer1997}
J.~Kuhn and C.~Rohrer.
\newblock 1997.
\newblock Approaching ambiguity in real-life sentences -- the application of an
  optimality theory-inspired constraint ranking in a large {LFG} grammar.
\newblock In {\em Proc. DGfS-CL, Heidelberg/FRG}.

\bibitem[\protect\citename{Kuhn \bgroup et al.\egroup
  }1998]{KuhnEckleRohrer1998}
J.~Kuhn, J.~Eckle, and C.~Rohrer.
\newblock 1998.
\newblock Lexicon acquisition with and for symbolic {NLP}-systems.
\newblock In {\em Proc. 1st Int'l Conf. on Language Resources and Evaluation},
  pages 89--95. Granada/ES.

\bibitem[\protect\citename{Lehmann and Oepen}1996]{Lehmann1996}
S.~Lehmann and S.~Oepen.
\newblock 1996.
\newblock Tsnlp - test suites for natural language processing.
\newblock In {\em Proc. 16th Int'l Conf. on Computational Linguistics}, pages
  711--716. Copenhagen/DK.

\bibitem[\protect\citename{Netter \bgroup et al.\egroup }1998]{Netter1998}
K.~Netter, S.~Armstrong, T.~Kiss, J.~Klein, and S.~Lehman.
\newblock 1998.
\newblock Diet - diagnostic and evaluation tools for nlp applications.
\newblock In {\em Proc. 1st Int'l Conf. on Language Resources and Evaluation},
  pages 573--579. Granada/Spain, 28-30 May 1998.

\bibitem[\protect\citename{Oepen and Flickinger}1998]{Oepen1998}
S.~Oepen and D.P. Flickinger.
\newblock 1998.
\newblock Towards systematic grammar profiling:test suite techn. 10 years afte.
\newblock {\em Journal of Computer Speech and Language}, 12:411--435.

\bibitem[\protect\citename{Rayner and Carter}1996]{Rayner1996}
M.~Rayner and D.~Carter.
\newblock 1996.
\newblock Fast parsing using pruning and grammar specialization.
\newblock In {\em Proc. 34th Annual Meeting of the ACL}. Santa Cruz, USA.

\bibitem[\protect\citename{Rayner and Samuelsson}1994]{Rayner1994}
M.~Rayner and C.~Samuelsson.
\newblock 1994.
\newblock Corpus-based grammar specialization for fast analysis.
\newblock In M.-S. Agn{\"{a}}s, H.~Alshawi, I.~Btrean, D.~Carter, and K.~Ceder,
  editors, {\em Spoken Language Translator: First-Year Report}, pages 41--54.
  Report CRC-043, Cambridge/UK: SRI International.

\bibitem[\protect\citename{Samuelsson}1994]{Samuelsson1994}
C.~Samuelsson.
\newblock 1994.
\newblock Grammar spezialization through entropy thresholds.
\newblock In {\em Proc. 32nd Annual Meeting of the ACL}.

\end{thebibliography}

\end{document}